# Explainable Artificial Intelligence and Machine Learning: A reality rooted perspective

Frank Emmert-Streib[1,2], Olli Yli-Harja[2], and Matthias Dehmer[3]

[1]Predictive Society and Data Analytics Lab, Faculty of Information Technology and Communication Sciences, Tampere University, Tampere, Finland *
[2]Institute of Biosciences and Medical Technology, Tampere University of Technology, Tampere, Finland
[3]Institute for Intelligent Production, Faculty for Management, University of Applied Sciences Upper Austria, Steyr Campus, 4040 Steyr, Austria

January 26, 2020

## Abstract

We are used to the availability of big data generated in nearly all fields of science as a consequence of technological progress. However, the analysis of such data possess vast challenges. One of these relates to the explainability of artificial intelligence (AI) or machine learning methods. Currently, many of such methods are non-transparent with respect to their working mechanism and for this reason are called black box models, most notably deep learning methods. However, it has been realized that this constitutes severe problems for a number of fields including the health sciences and criminal justice and arguments have been brought forward in favor of an explainable AI. In this paper, we do not assume the usual perspective presenting explainable AI as it should be, but rather we provide a discussion what explainable AI *can be*. The difference is that we do not present wishful thinking but reality grounded properties in relation to a scientific theory beyond physics.

## 1 Introduction

Artificial intelligence (AI) and machine learning (ML) have achieved great successes in a number of different learning tasks including image recognition and speech processing [1–3]. However, many of the best performing methods are too complex (abstract) prohibiting a straight forward explanation of the obtained results in simple words. The reason therefore is that such methods process high-dimensional input data in a non-linear and nested fashion to reach probabilistic decisions. This convolutes a clear view, e.g., on what information in the input vector is actually needed to arrive at certain decisions. As a result, such models are non-transparent or opaque and are typically regarded as *black box* models [4]. Importantly, not only deep neural networks (DNNs) are suffering from this shortcoming but also support vector machines (SVMs), random forests (RFs) or ensemble models (e.g. Adaboost) [5–8].

---

*frank.emmert-streib@tuni.fi

This black box character establishes problems for a number of fields. For instance, when making decisions in a hospital about the treatment of patients or at the court about the sentencing of a defendant, such decisions should be explainable [9, 10]. In an endeavor to address this issue the field explainable AI (XAI) has recently re-emerged [11]. While previous work in this area focused on specific problems of deep learning models, defining explainable AI or the taxonomy of XAI [12–14], our approach presents a different perspective as follows. First, instead of describing AI systems with desirable properties making them explainable, we present a reality rooted perspective showing what XAI can deliver. Put simply, instead of presenting explainable AI as it *should be* we show what explainable AI *can be*. Second, we derive thereof limitations of explainable AI. Such limitations may be undesirable but they are natural and unavoidable. Third, we discuss consequences of this for our way forward.

Our paper is organized as follows. In the next section, we briefly describe the current state of explainable AI. Then we present different perspectives on learning methods and discuss the definition of a scientific theory. This allows us to conclude some limitations of an even perfect versions of explainable AI. Finally, we discuss reasons for the confusion about explainable AI and present ways forward. The paper finishes with concluding remarks.

## 2 Current state of explainable AI

For our following discussion, it is important to know how explainable AI is currently defined. Put simply, one would like to have explanations of internal decisions within an AI system that lead to an external result (output). Such explanations should provide insight into the rationale the AI uses to draw conclusions [15].

A more specific definition of explainable AI was proposed in [16].

**Definition 1** (explainable AI). *1) produce more explainable models while maintaining a high level of learning performance (e.g., prediction accuracy), and 2) enable humans to understand, appropriately trust, and effectively manage the emerging generation of artificially intelligent partners.*

Another attempt of a definition of explainable AI is given in [13].

**Definition 2** (explainable AI). *Explainable Artificial Intelligence is a system that produces details or reasons to make its functioning clear or easy to understand.*

Furthermore, it is argued that the goals of explainable AI are trustworthiness, causality,

transferability, informativeness, confidence, fairness, accessibility, interactivity and privacy awareness [13, 17].

In general, there is agreement that an explainable AI system should not be opaque (or a black box) that hides its functional mechanism. Also, if a system is not opaque and one can even understand how inputs are mathematically mapped to outputs then the system is interpretable [15]. Taken together, this implies model transparency. The terms interpretability and explainability (and sometimes comprehensibility) are frequently used synonymously although the ML community seems to prefer the former while the AI community prefers the latter [14].

From a problem-oriented view, in [18] different types of interpretability, for instance perceptive interpretability, interpretability via mathematical structure, data-driven interpretability or interpretability by utilities, and explainability are discussed. They found that many journal papers in the ML and AI community are algorithm-centric, whereas in medicine risk and responsibility related issues are more prevalent. Similar discussions for different interpretations of interpretability can be found in [19].

Overall, at to this point one can conclude the following. First, there is no single definition of explainable AI available that would be generally accepted but many descriptions are overlapping with each other in the above discussed ways. Second, the characterizations of explainable AI state what XAI *should be*. That means they form desirable properties of such an AI system without deriving these from higher principles. Hence, these characterizations can be seen as wishful thinking. Before we can formulate reality rooted attainable goals of explainable AI, we need to discuss different perspectives on models and the general capabilities of a scientific theory.

## 3  Perspective of statistics

In statistics, one can distinguish between two main types of models. The first type, called inferential or explanatory model, provides a causal explanation of the data generation process whereas the second type, called predictive model, just produces forecasts [20, 21]. Certainly, an inferential model is more informative (i.e. theory-like, see below) than a predictive model because also an explanatory model can be used to make predictions but the predictive model does not provide (causal) explanations for such predictions. An examples for an explanatory model is a causal Bayesian network whereas a random forest is a prediction model. Due to the complementary capabilities of predictive and inferential models they are coexisting next to each other and each is useful in its own right.

The general problem for creating causal models from data is that their inference from observational data is very challenging requiring usually also experimental data (generated by perturbations of the system). Currently, most data in the health and social sciences are observational data ob-tained from merely observing the behavior of the system because performing experiments is either not straight forward or ethically prohibited.

## 4 Perspective of artificial intelligence

In [22] it was argued that explainable AI is not a new field but has been already recognized and discussed for expert systems in the 1980s. This is understandable considering that from about the 1950s to the 1980s the dominant paradigm of AI was symbolic artificial intelligence (SAI) and SAI used high-level and human-readable representations and rules for manipulating these, e.g., by using expert systems. Hence, not only the need for explainable systems has been realized but some AI systems could also accomplish near optimal explainable goals due to the nature of SAI.

With the renewed interest in neural networks in recent years in the form of deep neural networks (DNN), the question of explaining and interpreting models has been re-surfaced. One reason for this is that deep neural networks, in contrast to methods used for SAI, are not symbol-based but connectionist, i.e., they are learning (possibly high-dimensional) features from data and store them in the weights of the network [23, 24]. Although, more powerful in practice the price payed for this comes in form of a higher complexity of the representation used, which is no longer human-readable. Recently, it has been argued that AI systems should not only solve pattern recognition problems but provide causal models of the world that support explanation and understanding [25]. This demand connects directly to the statistics perspective because causal models are exploratory, see above.

After clarifying how explainable models are understood by different communities we take a step back to see what would be the ultimate goal achievable of an explainable AI. For this reason, we discuss the meaning of a theory.

## 5 What is a scientific theory?

In science, the formal definition of a theory is difficult but commonly it refers to a comprehensive explanation of a subfield of nature that is supported by a body of evidence [26–28]. In physics, the term theory is generally associated with a mathematical framework derived from a set of basic axioms or postulates which allows to generate experimentally testable predictions for such a subfield of physics. Typically these systems are highly idealized, in that the theories describe only certain aspects. Examples include classical field theory and quantum field theory. An important aspect of a theory is that it is falsifiable [29]. That means experiments can be conducted for testing the predictions made by a theory. As long as such experiments do not contradict the predictions a theory is accepted, otherwise rejected.

With respect to the stringency with which theories have been quantitatively tested, theories in physics, e.g., general relativity or quantum electrodynamics, are certainly what can be considered the best scientific theories. This implies that such theories provide answers to all questions that can be asked within the scope of the theory. Furthermore, the theory provides also an explanation of the obtained results. However, these explanations do not come in the form of a natural language, e.g., English, but are mathematical formulas. Hence, the mathematical formulas need to be interpreted by means of a natural language as good as possible. This may seem as a minor issue but the severity of this may be exemplified by the interpretation of quantum mechanics because so far there is no generally accepted interpretation, although the Copenhagen interpretation is the most popular one [30]. Interestingly, the reason for this is ascribed to personal philosophical prejudice [30]. The latter point hints to the incompleteness of any natural language in interpreting a mathematical formalism of quantum mechanics. For completeness, we would like to mention that even in physics not everything is covered by the existing theories because so far there is no theory unifying gravity and quantum mechanics [31].

# 6 Expected limitations of an explainable AI

From this discussion, we can conclude some limitations even a perfect version of an explainable AI will have. Considering that essentially all applications of AI and ML are beyond physics, e.g., in biology, medicine and health care, industrial production or human behavior, one cannot expect to have a simpler theory for any of these fields than what we have for physics. Hence, even the interpretability of such a theory is expected to be more problematic than an interpretation of, e.g., quantum mechanics.

In order to make this point more clear let's consider a specific example. Suppose a theory of cancer would be known, in the sense of a theory in physics discussed above. Then this cancer theory would be highly mathematical in nature which would not permit a simple one-to-one interpretation in any natural language. Hence, only highly trained theoretical cancer mathematicians (in analogy to theoretical physicists) would be able to derive and interpret meaningful statements from this theory. Regardless of the potential success of such a cancer theory, this implies that medical doctors - not trained as theoretical cancer mathematicians - could not understand nor interpret such a theory properly and, hence, from their perspective the cancer theory would appear opaque or non-explainable. Without our discussion, such a result would appear undesirable and unacceptable, however, given the context we provided above this appears unavoidable and natural. A similar discussion can be provided for any other field than cancer showing that even a perfect version of an explainable AI theory would not be interpretable or explainable for managers or general practitioners for natural reasons.

The next optimistic but less ambitious assumption would be to suppose AI could provide a *description* for fields outside of physics. Interestingly, physicists realized already decades ago that such an expansion of a theory beyond the boundaries of physics is very challenging. For instance, severe problems encountered are due to the arising of emergent properties and non-equilibrium systems [32,33]. For this reason, phenomena outside of physics are usually addressed by *"approaches"* collectively named as complex adaptive systems (CAS) [34,35]. We used intentionally the word *"approaches"* and not *"theories"* because the used models and the obtained descriptions are qualitatively very different thereof. Whereas it is unquestionable that valuable insights have been gained into complex phenomena from economy, society and sociology [36,37] a theory for such fields is still absent and currently not in sight. Hence, it seems fair to conclude that even the most powerful AI system of CAS would be far from being a theory and for this reason lack explanatory capabilities. However, this translates directly into a principle incompleteness of questions such an AI system could answer and reduces further what could be delivered by an explainable AI.

Finally, we come to the most realistic view on an AI system which views its purpose as a system to analyze data. Depending on the AI system and the available data it is clear that the understanding that can be obtained from such a system is even further limited in the answers that can be provided as well as in the level of explanations it can give. This is also true if the AI system would be based on a perfect method because the data represent only an incomplete sample of all possible data and is as such inherently limited in the explanations it can provide translating in an unavoidable uncertainty of statements about the population it studies.

In Fig. 1, we summarize the above discussion graphically. The shown coordinate systems indicate the qualitative relation between the influence of the distance from a theory and the comprehensiveness of the description (left) and the influence of the sample size on the uncertainty of statements or explanations about the population (right). The yellow arrow on the left indicates the distance from physical theories which corresponds also to the x-axis of the left coordinate system. A similar meaning has the purple arrow on the right for the diameter of the random sample and x-axis of the right coordinate system

## 7 Reasons for the confusion

It is interesting to ask how utopian, idealistic requirement for an explainable AI, as discussed in Section 2 above, could be demanded when reality looks quite differently. We think the reason for this is twofold. First, statistical models and machine learning methods have been introduced due to the lack of general theories outside of physics because these allow a quantitative analysis studying experimental evidence. Hence, even in this unsatisfactory situation, systematic and quantitative approaches are available for extending our knowledge of complex phenomena. Second, in recent years we have been experiencing a data surge which gives the impression that every research question should start with "data". This led even to the establishment of a new field called data science [38]. Taken together, this may have given the impression that AI and ML methods are more powerful than physical theories because they can (approximately) reach areas - due to the availability of methods and data - that are blocked for physics. However, these methods are not intended as theories but merely as practical instruments to deal with complex phenomena. Importantly, the questions that can be answered need to have answers that can be found within the data. Every lack of such qualitative data, e.g., due to a limited sample size, translates directly into a lack of answers and is for this reason an inherent uncertainty of any AI system.

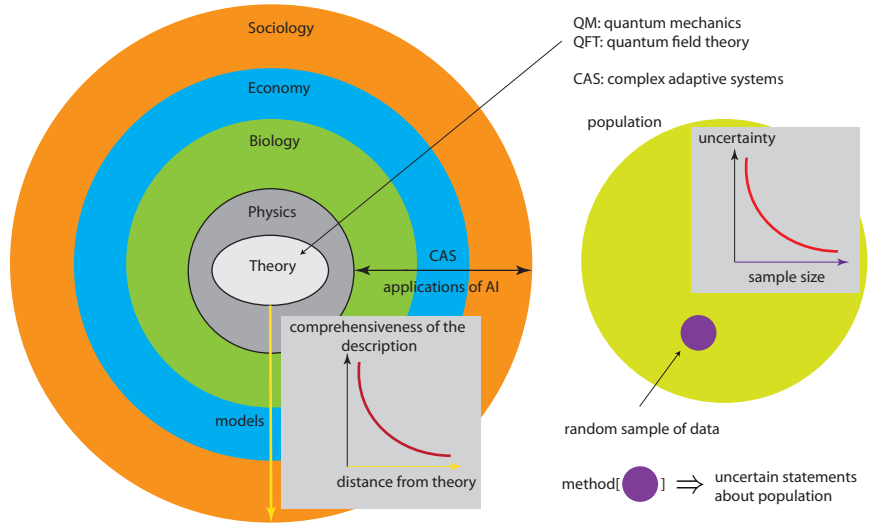

Figure 1: An overview describing the limitations of explainable AI with respect to attainable goals. Left: Different scientific fields are arranged according to their increasing complexity [32] starting from the best (most comprehensive) theories of physics in the center. The further the distance from these theories the less comprehensive are the models describing subjects of CAS (left coordinate system). Right: Any AI system analyzes a random sample of data drawn from a population. One source of uncertainty is provided by the sample size of the data (right coordinate system) that translates directly into uncertain statements about the population.

## 8 Discussion

Having realized the limitations of explainable AI with respect to attainable goals, what does this mean for the way to go forward? In the following, we present a discussion of practical remedies that are based on the insights gained in the first part of our paper. We want to emphasize that these remedies do not bring us back to the delusional view of an idealized explainable AI but provide means to realistically formulate achievable goals.

*An AI system may not need to be explainable in a natural language as long as its generalization error does not exceed an acceptable level.* An example for such a practice are nuclear power plants.

Since nuclear power plants are based on physical laws (see the discussion of quantum mechanics above) the functioning mechanisms of such plants are not explainable to members of the regulatory management or politicians responsible for governing energy politics. Nevertheless, nuclear power plants have been operated since many decades contributing to our energy supply. In a similar way, one could envision an operational policy for medicine, e.g., utilizing deep learning methods.

*Deep learning may not be necessary to analyze a given data set.* Nowadays, many people are using deep learning methods because they seem to think such methods are needed without exploring alternative approaches. However, in this way problems regarding the interpretability and explainability of the results are encountered. While it may be possible that future deep learning methods may be less opaque, if currently available methods solve the same problems in a satisfactory way and do not suffer from such limitations, e.g., decision trees, they should be preferred and used.

*The similarity (or difference) of the predictiveness of AI systems needs to be quantified in an explainable way.* This point is related to the previous one because if one can quantitatively assesses the similarity (or the difference) between two AI systems one can compare an explainable with a non-explainable AI system to evaluate one benefit over the other. Hence, even if an explainable AI system does not fully solve a given problem, e.g., compared to a deep learning approach, it may be sufficient to use. Importantly, even when an AI system itself may not be explainable the comparison of different systems can be understandable. This way the lack of an explainability maybe compensatable. A challenge of such an approach is that a quantification should not only be based on prediction errors [39] but an assessment of risk and utility [40]. This latter issue is clear in a clinical context.

*Partial insights into the interpretability and explainability of AI systems should be developed.* It may not be feasible to convert deep learning models into fully transparent systems, however, this may be achievable in part. For instance, certain aspects of an analysis could be understandable which are integrated to achieve the complete model. Given the fact that in general data science problems present themselves as a process [41] there should be ample opportunity to identify subproblems deemable as an explainable AI.

## 9 Conclusions

In this paper, we shed some light on the current state of explainable AI and derived limitations with respect to attainable goals from clarifying different perspectives and the capabilities of well tested physical theories. The main results can be summarized as follows:

1. An AI system does not constitute a theory but an instrument (a model) to analyze data. ⇒ The comprehensiveness and the explainability of an even perfect AI system are inherently limited by the random sample of data used.

2. The more comprehensive (i.e. theory-like) an AI system becomes in predicting CAS the more complex becomes its underlying mathematics. $\Rightarrow$ There is no simple one-to-one translation into a natural language to explain the results or the working mechanism.

3. The most powerful but opaque AI systems (e.g. deep learning) should not be preferred and applied by default but a comparison to alternative explainable methods should be conducted and differences should be quantified. $\Rightarrow$ An explainable and quantifiable reason can be derived by integrating prediction error, risk and utility for weighing the pros and cons for each model.

We hope our results can contribute to formulating realistic goals for an explainable AI that are also attainable [42].

## Conflict of Interest Statement

The authors declare that the research was conducted in the absence of any commercial or financial relationships that could be construed as a potential conflict of interest.

## Author Contributions

All authors contributed to all aspects of the preparation and the writing of the manuscript.

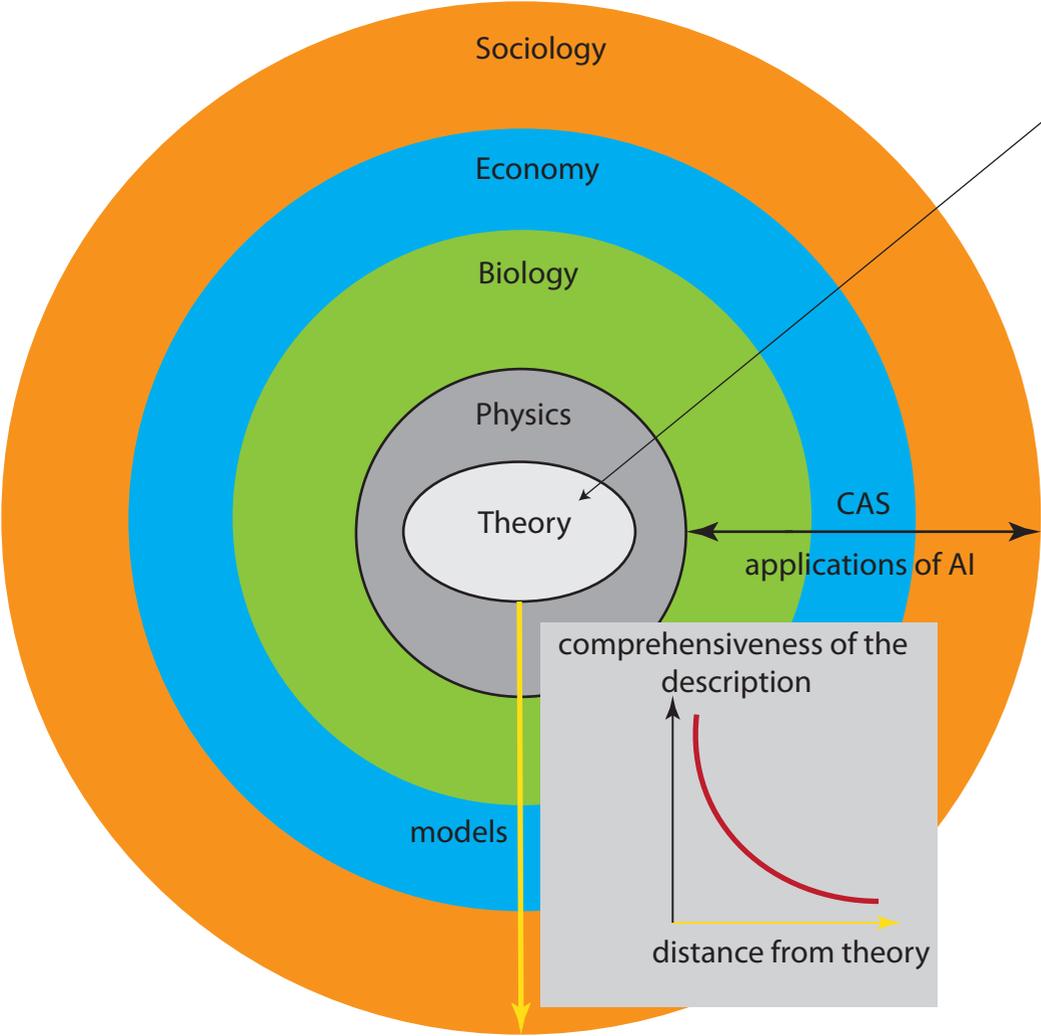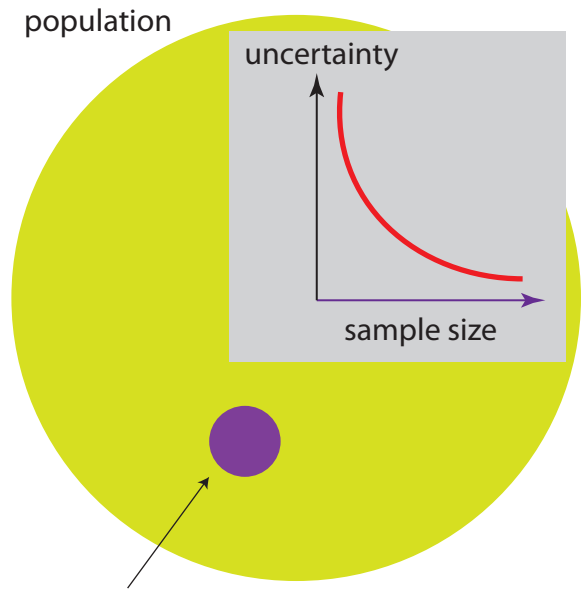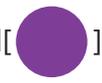